\documentclass{article}
\usepackage{ijcai20}

\usepackage{times}

\usepackage{soul}
\usepackage{url}
\usepackage[hidelinks]{hyperref}
\usepackage[utf8]{inputenc}
\usepackage[small]{caption}
\usepackage{graphicx}
\usepackage{amsmath}
\usepackage{amsthm}
\usepackage{booktabs}
\usepackage{subfigure}
\usepackage{algorithm}
\usepackage{algorithmic} 

\newtheorem{definition}{Definition}

\usepackage[normalem]{ulem}
\usepackage{color}

\newcounter{ToDo}
\newcounter{gaocomm}
\newcounter{Note}
\definecolor{blue-violet}{rgb}{0.54, 0.17, 0.89}
\definecolor{mygreen}{rgb}{0.0, 0.5, 0.0}
\definecolor{awesome}{rgb}{1.0, 0.13, 0.32}
\definecolor{bostonuniversityred}{rgb}{1.0, 0.0, 0.0}

\title{Differential Evolution with Individuals Redistribution \\for Real Parameter Single Objective Optimization}

\author{
Chengjun Li
\And
Yang Li\\
\affiliations
School of Computer Science, China University of Geosciences, Wuhan, China
\emails
\{chengjun\_li, liyang\_cs\}@cug.edu.cn
}

\begin{document}

\maketitle

\begin{abstract}
Differential Evolution (DE) is quite powerful for real parameter single objective optimization.
However, the ability of extending or changing search area when falling into a local optimum is still required to be developed in DE for accommodating extremely complicated fitness landscapes with a huge number of local optima. 
We propose a new flow of DE, termed DE with individuals redistribution, in which a process of individuals redistribution will be called when progress on fitness is low for generations. 
In such a process, mutation and crossover are standardized, while trial vectors are all kept in selection. 
Once diversity exceeds a predetermined threshold, our opposition replacement is executed, 
then algorithm behavior returns to original mode.
In our experiments based on two benchmark test suites, we apply individuals redistribution in ten DE algorithms.
Versions of the ten DE algorithms based on individuals redistribution are compared with not only original version but also version based on complete restart, where individuals redistribution and complete restart are based on the same entry criterion.
Experimental results indicate that, for most of the DE algorithms, version based on individuals redistribution performs better than both original version and version based on complete restart.
\end{abstract}

\section{Introduction}

Real parameter single objective optimization is one of the active research fields in artificial intelligence and the basis of more complex optimization such as multi-objective optimization, constrained optimization, etc.~\cite{brest2017single}. 
So far, a variety of population-based metaheuristics have been proposed for this type of optimization.
In the series of Congress of Evolutionary Computation (CEC), at least five competitions among population-based metaheuristics on real parameter single objective optimization have been held since 2013.
We briefly introduce the winner of each of five competitions as follows.
\begin{itemize}
\item 
NBIPOP-aCMA-ES~\cite{loshchilov2013cma} is a variant of Covariance Matrix Adaptation Evolution Strategy (CMA-ES)~\cite{hansen2003reducing} with a restart scheme including two regimes; 
\item L-SHADE~\cite{tanabe2014improving} is a Differential Evolution (DE)~\cite{storn1997differential} algorithm extended from SHADE~\cite{tanabe2013success} with linear population size reduction.
\item
L-SHADE-EpSin~\cite{awad2016ensemble} is a variant of L-SHADE with adaptive parameter
settings and a local search based on Gaussian walks.
\item
UMOEAs-\uppercase\expandafter{\romannumeral2}~\cite{elsayed2016testing} employs two subpopulations evolved by an ensemble DE and CMA-ES, respectively.
\item
EBOwithCMAR~\cite{kumar2017improving} is a self-adaptive butterfly optimizer with success history based adaption, linear population size reduction, and covariance matrix adapted retreat phase; and \item
HS-ES~\cite{zhang2018hybrid} is CMA-ES with univariate sampling.
\end{itemize}
Note that L-SHADE-EpSin and UMOEA-\uppercase\expandafter{\romannumeral2} are joint winners in 2016.  Among the six winners, L-SHADE and L-SHADE-EpSin are based on DE, while NBIPOP-aCMA-ES and HS-ES are based on CMA-ES.
Moreover, UMOEAs-\uppercase\expandafter{\romannumeral2} is based on both DE and CMA-ES. 
In~\cite{vskvorc2019cec}, the six winners
are compared.
The comparison shows that L-SHADE-EpSin, UMOEA-\uppercase\expandafter{\romannumeral2}, and HS-ES are the three top performers.

In brief, both DE and CMA-ES perform powerfully for real parameter single objective optimization and deserve to be further studied.
In this paper, we choose to scrutinize DE.

In DE, population, the set of tentative solutions, i.e. individuals or target vectors, is driven by the operators - mutation, crossover, and selection.
In generation $g$, mutant vectors $\vec v_{i,g} (i=1,2,...,NP)$,  where $NP$ denotes population size, are produced based on target vectors $\vec x_{i,g}$ by mutation. 
After mutation, trial vectors $\vec u_{i,g}$ are generated based on both $\vec x_{i,g}$ and $\vec v_{i,g}$ by crossover.
Thus, in DE, crossover and mutation together are specified as trial vector generation strategy.
For selection, $\vec x_{i,g+1}$ is selected from $\vec x_{i,g}$ or $\vec u_{i,g}$ based on their fitness computed by function evaluation.
\paragraph{Motivation.}
Either benchmark test functions or real-world problems of real parameter single objective optimization often have very complex fitness landscapes.
It is very common that there are a huge number of local optima in fitness landscapes.
The above fact is a challenge to algorithms.
During an execution of DE for real parameter single objective optimization, individuals tend to get closer and closer in distance under the control of the operators.
That is, diversity of population goes lower and lower.
Thus, search area decided by individuals distribution becomes smaller and smaller. 
Then, all or most individuals may enter the neighbourhood of an optimum.
After that, although the found optimum may be just a local optimum, no solution better than the optimum can be found any more because the optimum is the best solution in the current search area. 
Therefore, extending and changing search area when execution has fallen into a local optimum is required to keep long-term search ability of DE for real parameter single objective optimization.


Six population-based metaheuristics from the CEC competitions including five winners are executed by us for much larger Maximum Number of Function Evaluations $MFES$ than the value used in existing competitions and papers.
We find that these algorithms for real parameter single objective optimization can hardly make progress in the latter stage of execution.
That is, the similar tendency of falling into local optimum widely exists in different types of population-based metaheuristics for real parameter single objective optimization.
Considering the leading position of DE in the field of real parameter single objective optimization, our idea of improving the ability of DE will contribute to the whole field.

\begin{figure}[t]
\centering
  \subfigure[The initial state]{\includegraphics[width=1.1in]{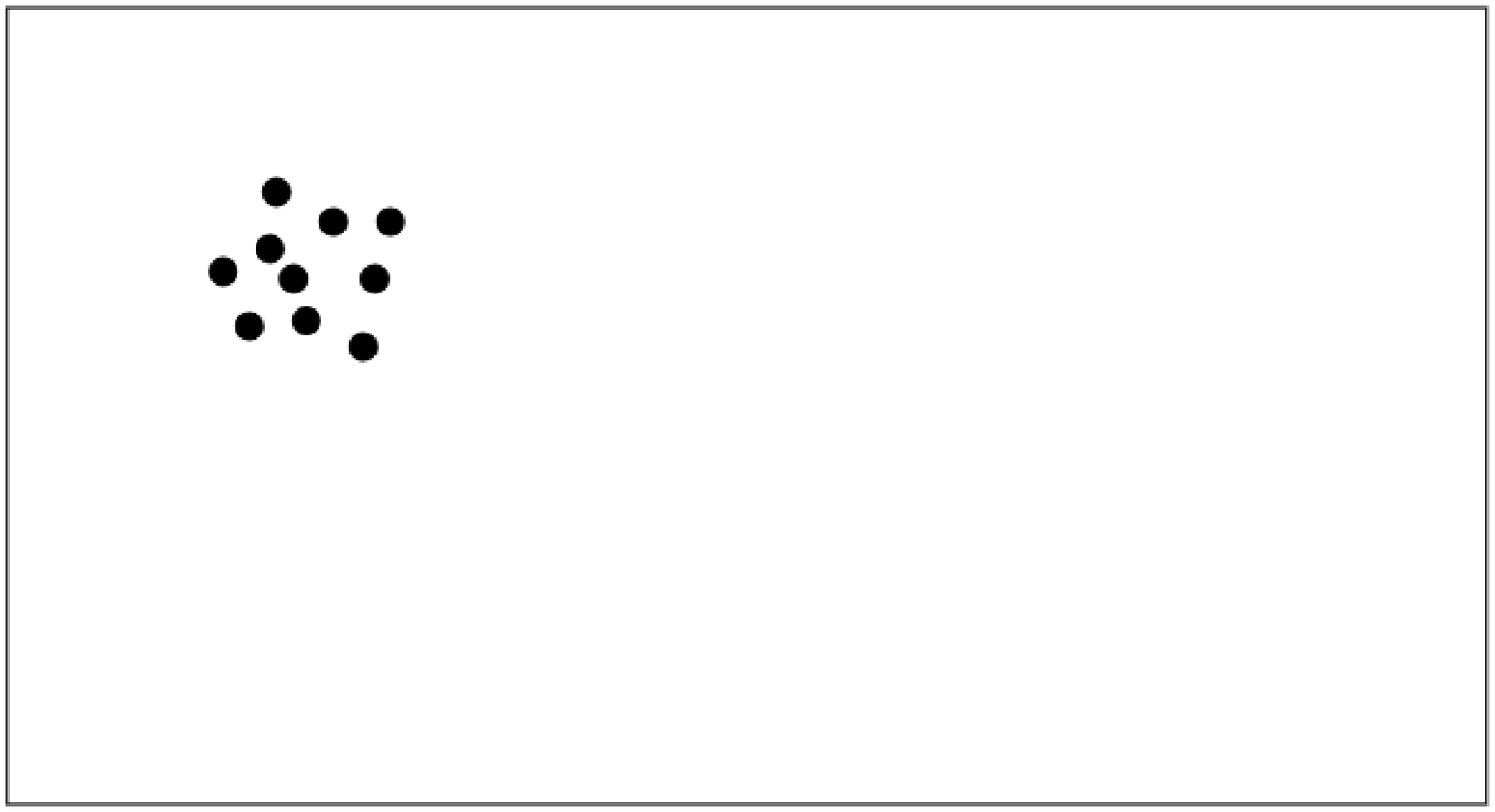}}
  \subfigure[The first step]{\includegraphics[width=1.1in]{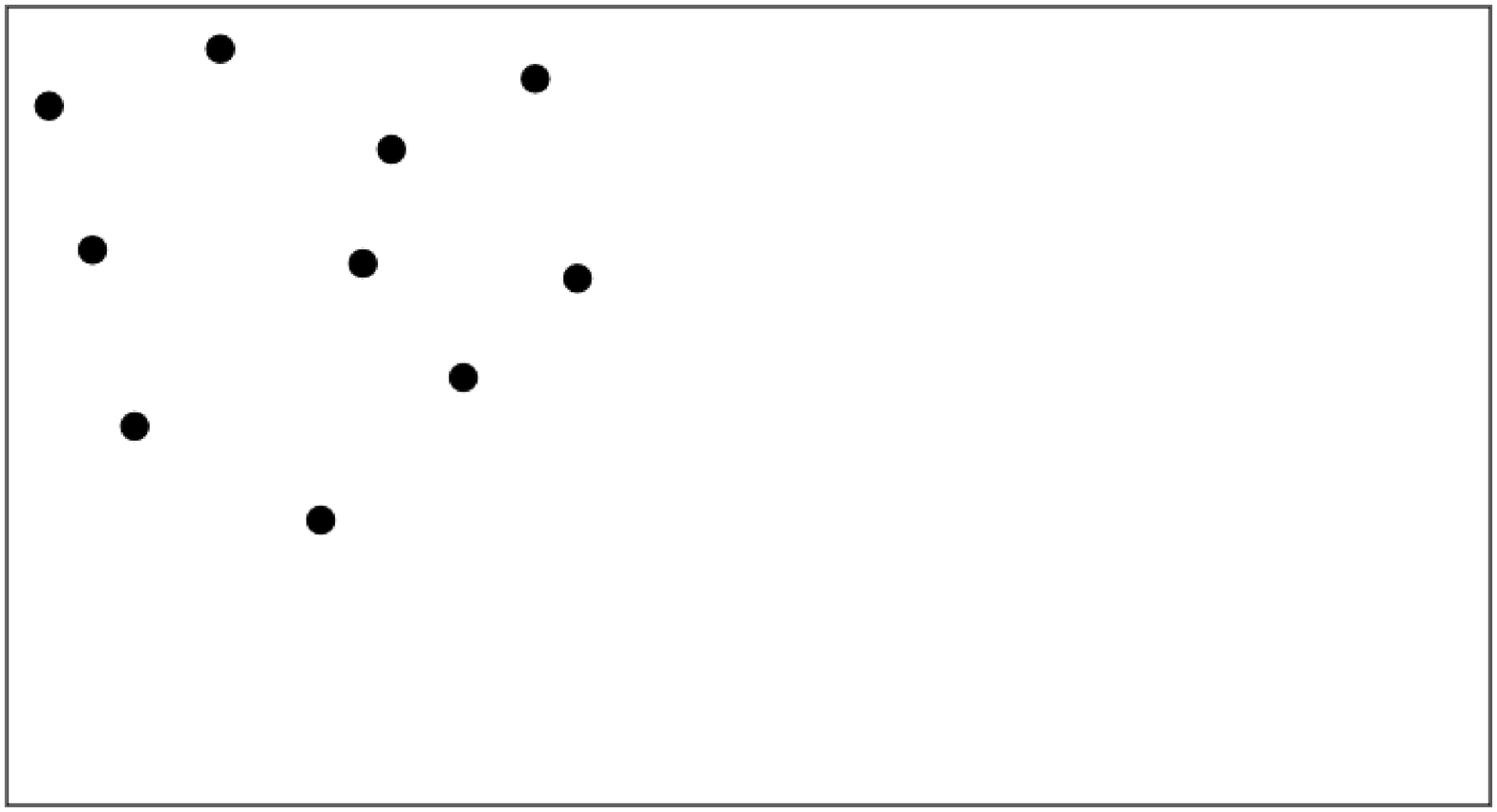}}
  \subfigure[The second step]{\includegraphics[width=1.1in]{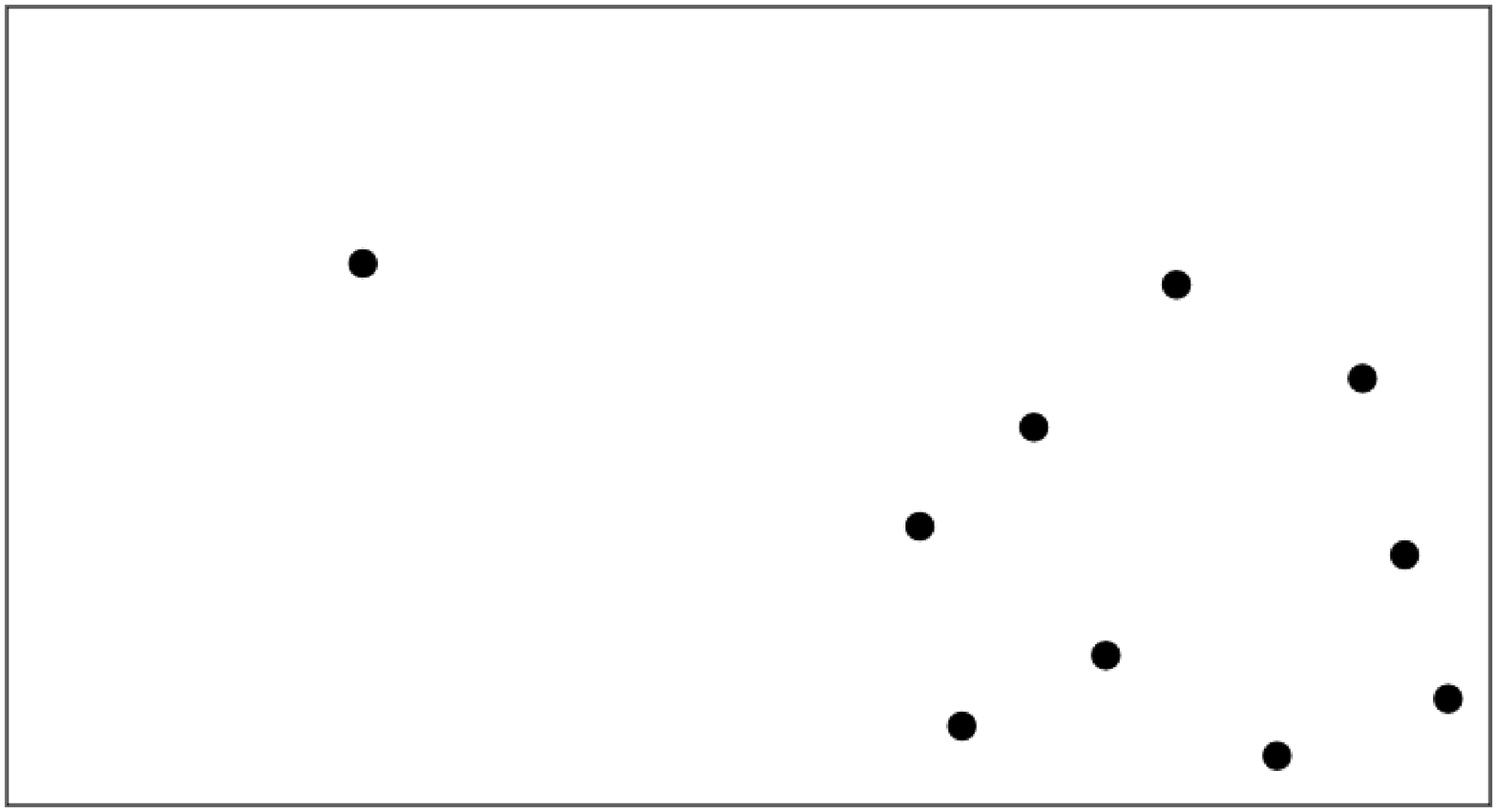}}
\caption{Sketch of process of individuals redistribution in two-dimensional search space}
\label{Fig:01}
\end{figure}

In this paper, we propose to equip DE with individuals redistribution.
In the scheme, a process of individuals redistribution starts given that progress on fitness is continuously lower than a predetermined degree for a certain number of generations.
In such a process, mutation is changed to DE/rand/1
with $F=1.0$, while crossover is set to binomial crossover with $CR=0.5$.
Meanwhile, trial-vector-kept are used for selection.
The changed operators may make that diversity goes higher and higher.
Once diversity exceeds a predetermined threshold or the number of generations with
changed operators exceeds a predetermined value, a part of target vectors are replaced by their opposite vector in search space.
The definition of opposite vector in search space can be seen in our methodology.
In Figure~\ref{Fig:01}, we give the sketch of individuals redistribution.
After individuals redistribution, algorithm behavior returns to original mode.

Our scheme can be widely applied in different DE algorithms.
In our experiments, ten DE algorithms for real-parameter single objective optimization
are employed, as listed below,
\begin{itemize}
\item L-SHADE~\cite{tanabe2014improving}
is extended from SHADE~\cite{tanabe2013success} - a variant of JADE~\cite{zhang2009jade}.
\item CoBiDE~\cite{wang2014diff} is based on covariance matrix learning and parameters adaption.
\item L-SHADE-EpSin~\cite{awad2016ensemble}
is extended from L-SHADE.
\item SaDE/Mexp~\cite{qiu2016multiple} is an improved version of SaDE~\cite{qin2005self} with 
multiple exponential crossover.
\item MPEDE~\cite{wu2016differential} is based on parameters adaption and three subpopuations controlled by different
mutation strategies.
\item ETI-JADE~\cite{du2017differential} is an improved version of JADE with
Event-Triggered Impulsive (ETI) scheme.

\item L-SHADE-RSP~\cite{stanovov2018lshade} ranks second in the CEC 2018 competition.
Based on L-SHADE, the algorithm employs rank-based mutation scheme in which better target vector has larger possibility to be chosen in the secondary differential item.
\item EDEV~\cite{wu2018ensemble} is based on three competing subpopulations controlled by JADE, CoDE~\cite{wang2011differential}, and EPSDE~\cite{mallipeddi2011differential}.
\item MLCC-SI~\cite{zhang2019multi} is a multi-layer competitive-cooperative framework based on SHADE and IDE~\cite{tang2015differential}.
\item NDE~\cite{tian2019differential} is with neighborhood-based mutation strategy, adaptive evolution mechanism, and linear population size reduction.
\end{itemize}

The state-of-the-art DE algorithms - SaDE, JADE, CoDE, EPSDE, CoBiDE, and IDE - are involved in our experiments directly or indirectly, and the up-to-date DE algorithms - MLCC-SI and NDE - as well.
Further, two winners and a good performer in competitions - L-SHADE, L-SHADE-EpSin, and L-SHADE-RSP - are studied in this paper.
In fact, L-SHADE-EpSin is also one of the top performers in all the competitions.


The CEC 2014 and CEC 2017 benchmark test suites are both used in our experiments.
After large $MFES$, results of version based on our individuals redistribution are compared with not only original version but also version based on complete restart.
Results show that version based on complete restart performs better than original version.
More importantly, for most of the DE algorithms, version with individuals redistribution performs better than not only original version but also version with complete restart.

The major contribution of this paper is summarized as
\begin{itemize}
\item showing that, for real parameter single objective optimization, population-based metaheuristics including DE tend to fall into local optimum in many occasions;
\item proposing a new flow of DE - 
 DE with individuals redistribution - for extending and changing search area to jump out of local optimum; and
\item validating general applicability and effectiveness of the new flow by experiments based on ten DE algorithms and two benchmark test suites.
\end{itemize}

\section{Data Analysis}
Here we run six metaheuristics selected from the competitions - L-SHADE, L-SHADE-EpSin, UMOEAs-\uppercase\expandafter{\romannumeral2},
EBOwithCMAR, jSO \cite{brest2017single}, and HS-ES.
Except that jSO ranks second in
the CEC 2017 competition, each of the other five algorithms is the winner in different competitions.
These algorithms are executed for the CEC 2017 benchmark test suite.
In the suite, hybrid functions, F11-F20, and composition
functions, F21-F30, have much more local optima in fitness landscapes than unimodal functions, F1-F3, and simple
multimodal functions, F4-F10.
$MFES$ is set $3.0E+06$ for all the algorithms.
The value is 10 times as much as the value widely used in competitions and papers.
Besides, all the other parameters are taken the original value in literature.

According to results, global optimum of 21 functions - F5, F11-F30 is never found.
That is, it is hard for the six algorithms to obtain global optimum of F11-F30, the functions with a huge number of local optima in fitness landscapes.
For F11-F30, we plot the averages of the best error after the first generation and at 100 intervals of all the DE algorithms.
We find that, for the functions, all the algorithms can hardly make progress in the latter stage of execution.
In brief, even winner or good performer in the competitions among population-based metaheuristics for real parameter single objective optimization do fall into local optimum in many occasions.

The aforementioned data analysis shows that the tendency of falling into local optimum widely exists in different types of metaheuristics for real parameter single objective optimization including DE.
Now that DE performs well with possibility of falling into local optimum, a new flow of DE for extending and changing search area to leave local optimum is meaningful in the field of real parameter single objective optimization.   

\section{Related Work}
Firstly, recent methods which can be widely used in DE algorithms for improving solutions are briefly summarized here.
In multiple exponential crossover~\cite{qiu2016multiple}, multiple segments of individuals are exchanged.
SaDE/Mexp is based on multiple exponential crossover.
Under the control of ETI scheme~\cite{du2017differential}, stabilizing impulses and destabilizing impulses are executed based on event-triggered mechanism.
In fact, destabilizing impulses serve as a partial restart strategy.
ETI-JADE is based on ETI scheme.
Adaptive social learning strategy~\cite{cai2018social} can extract the neighborhood relationship information of individuals.
Multi-topology-based DE~\cite{sun2018differential} includes multiple population topologies, individual-dependent adaptive topology selection scheme, and topology-dependent mutation strategy.

Moreover, ensemble DE algorithms - DE algorithms with more than one combinations of operators - are also related to our work because, to apply our scheme, a combination of the changed operators need be added in DE algorithms.
In MPEDE, mutation is set differently in subpopulations.
In LSAOS-DE~\cite{sallam2017landscape}, each combination controls a subpopulation in a part of generations, whlie the combination showing the best performance seizes the whole population in the other generations.
According to \cite{cui2018adaptive}, two trial vector generation strategies both control the whole population in a part of generations, while 
population is controlled by only one of the two strategies in other generations. 
EDEV employs three state-of-the-art DE algorithms to control three subpopulations, respectively.
MLCC framework~\cite{zhang2019multi}
implements a parallel structure with the entire population simultaneously monitored by multiple DEs assigned to different layers. 
Thus, individuals are processed in different layers.
MLCC-SI is based on MLCC framework.


Among the DE algorithms listed in related work, 
SaDE/Mexp, MPEDE, ETI-JADE, EDEV, and MLCC-SI are involved in our experiments.
Thus, our scheme can be validated even based on related work.

\section{Methodology}
We propose DE with individuals redistribution - a new flow of DE - to improve DE at the aspect of jumping out of local optimum for keeping search ability in long run. 
As the name implies, individuals redistribution makes the distribution of individuals significantly change.
Details of our scheme are given as below.

Only if improvement on the best fitness is lower than a given threshold $T_{IR}$ for a given number of contiguous generations $G_N$, a process of individuals redistribution is called.
In this way, individuals redistribution is always executed when making progress is difficult, or, in other words, when the current run falls into a local optimum.
As soon as that the current best fitness becomes the best fitness in the current run, $G_N$ is doubled based on the given value to provide more generations for obtaining an ever-best solution.

In DE algorithms, operators are designed for convergence.
However, for individuals redistribution, original operators are replaced by the changed operators to diversify population.
For any DE algorithm, the changed operators are set same as below.
Firstly, original mutation strategy is changed to DE/rand/1 with scaling factor $F=1.0$ as shown in Equation~\eqref{eq1} 
\begin{equation}
\vec{v}_{i,g}=\vec{x}_{i,g}+1.0\cdot(\vec{x}_{r1,g}-\vec{x}_{r2,g})
\label{eq1}
\end{equation}
where $r1\in \{1,2,...,NP\}$, $r2\in \{1,2,...,NP\}$, and $r1\neq r2\neq i$.
Then, crossover is set to binomial crossover with $CR=0.5$.
Let $\vec{u}_{i,g}=(u_{1,i,g},u_{2,i,g},...,u_{n,i,g})$, where $n$ denotes the number of dimensions.
Our crossover method can be expressed by Equation~\eqref{eq2}.
\begin{equation}
u_{j,i,g}=\left\{
\begin{aligned}
v_{j,i,g} & , & \text{if }  \text{rand}(0,1)\leq 0.5 \\
x_{j,i,g} & , & \text{otherwise}.
\end{aligned}
\right.
\label{eq2}
\end{equation}
Further, trial-vector-kept, which means keeping all trial vectors, is used as selection in this stage.
In short, for individuals redistribution, two aspects are changed in algorithm.
On one hand, trial vector generation strategy is standardized.
More importantly, fitness is not considered in selection to remove selective pressure.
Thus, diversity may show an up trend.
The changed operators are executed in contiguous generations until diversity reaches a given value, $T_{DIV}$.

For real parameter single objective optimization, diversity can be denoted as the average of distance from individual to population center.
The definition of population center is given below.
\begin{definition}[Population center]
Let $P_g=\{\vec x_{1,g},\vec x_{2,g},...,\vec x_{NP,g}\}$ be the $g$th generation of population, 
where, $\vec{x}_{i,g}=(x_{1,i,g},x_{2,i,g},...,x_{n,i,g})(i=1,2,...,NP)$.
Let $m_j$ be the median of $x_{j,1,g}$,  $x_{j,2,g}$ , ..., $x_{j,NP,g}$ where $j=1,2,...,n$.
The center of $P_g$ is $c_g=(m_1,m_2,...,m_n)$
\end{definition}
Here, based on normalized Manhattan distance, diversity is calculated according to Equation~\eqref{eq3}, 
\begin{equation}
div_g=\frac{\sum_{i=1}^{NP}\sum_{j=1}^n\frac{|x_{j,i,g}-m_j|}{up_j-low_j}}{NP} 
\label{eq3}
\end{equation}
where $up_j$ and $low_j$ are upper limit and lower limit of the $j$th dimension, respectively.


After diversity reaches $T_{DIV}$, a portion of target vectors, which are randomly chosen, are replaced by their opposite vector in search space.
This step is called opposition replacement by us.
Then, evolution goes back to original mode.
The definition of opposite vector in search space is given below.

\begin{definition}[Opposite vector in search space]
Let $\vec x_{i,g}=(x_{1,i,g},x_{2,i,g},...,x_{n,i,g})$ be a target vector.
The opposite vector of $\vec x_{i,g}$ - $\vec x^o_{i,g}$,
is completely defined by its components 
\begin{equation}
x^o_{j,i,g}=up_j+low_j-x_{j,i,g}
\end{equation}
where $j\in\{1,2,...,n\})$ and $x_{j,i,g}\in[up_j,low_j]$.
\end{definition}


According to our scheme, once it is difficult to make progress in fitness any more, population is diversified gradually and slowly until diversity rises to $T_{DIV}$. 
Then, opposition replacement is done to make a portion of target vectors become far from the gathering position before individuals redistribution.
Meanwhile, the other target vectors still locate in the vicinity of the position.
After that, target vectors delimit a new search area.
The new search area is larger than the search area before individuals redistribution and has very limited overlap ratio with the old one.
It need be emphasized that our scheme preserves no elite. 

In Algorithm~\ref{alg00}, the pseudo-code of DE with individuals redistribution is given.
\begin{algorithm}[!h]\footnotesize
\caption{The pseudo-code of DE with individuals redistribution}
\label{alg00}
\textbf{Input}: $NP$, population size; $MFES$, the maximum number of function evaluations; $G_N$, the maximum number of sequential generations with no significant improvement in best fitness;  $T_{IR}$, threshold for the improvement ratio; $T_{DIV}$, threshold for diversity; $T_{GEN}$, threshold for generation;
$R$, the proportion of  opposition replacement\\
\textbf{Parameter}: $f_{eb}$, the best fitness from the generation after the end of the previous process of individuals redistribution to the previous generation; $f'_{eb}$, the best fitness from the initial generation to the previous generation;\\
\textbf{Output}: $S$, the best fitness found in whole run\\
\begin{algorithmic}[1]
\STATE Initialize and evaluate $\vec x_{i,0}$ $(i=1,2,...,NP)$
\STATE $FES=NP$, $g_n=0$, $g=0$, $b=0$, 
\WHILE{$FES<=MFES$}
 \STATE $applied\_G_N=G_N$
 \IF{$b==0$}
  \IF{$f(\vec x_{best,g})>=f_{eb}$ OR $(f(\vec x_{best,g})<f_{eb}$ AND $\frac{f_{eb}-f(\vec x_{best,g})}{f_{eb}}<T_{IR})$}
   \STATE{$g_n=g_n+1$}
  \ELSE 
   \STATE{$g_n=0$}
  \ENDIF
  \IF{$f_{eb}==f'_{eb}$}
   \STATE {$applied\_G_N=2\cdot G_N$}
  \ENDIF
  \IF{$g_n==applied\_G_N$}
   \STATE{$b=1$, $g_n=0$, $f_{eb}=real\_max$, $g_c=0$}
  \ENDIF
 \ENDIF
 \IF{$b==0$}
  \STATE{Execute original mutation and crossover}
  \STATE {Evaluate trial vectors to execute original selection}
  \STATE {$FES=FES+NP$, $g=g+1$}  
  \ELSE
   \STATE{Compute diversity, $div$}
   \STATE{$g_c=g_c+1$}
   \STATE {Execute the changed mutation and crossover}
  \IF{$div>T_{DIV}$ OR $g_c>T_{GEN}$}
\STATE Opposition replacement is done on $R\cdot NP$ individuals\STATE {Evaluate trial vectors}
  \STATE {$FES=FES+NP$, $g=g+1$, $b=0$}
 \ENDIF 
 \STATE{Execute changed selection}
 \ENDIF
\ENDWHILE
\STATE {Report $S$}
\end{algorithmic}
\end{algorithm}

$b=0$ means original mode, while $b=1$ means generations for individuals redistribution. 
It can be seen from Algorithm~\ref{alg00} that, if diversity is already higher than $T_{DIV}$ before the start of individuals redistribution, just one generation with the changed operators is executed.
Moreover, beside the main criterion - diversity higher than $T_{DIV}$, another criterion - the number of contiguous generations for the changed operators exceeding our predetermined limit - is also used for switching from the changed operators to original ones. 
We set the secondary criterion because, although it has a very high probability that diversity is increased beyond $T_{DIV}$ under the control of the changed operators after generations, very few counter examples can be found.

Neither the changed operators nor opposition replacement is not based on fitness.
Thus, during a process of individuals redistribution, function evaluations are required just after opposition replacement for the following old mode of evolution.
That is, only $NP$ times of function evaluations are needed in a process of individuals redistribution.
Nevertheless, the higher $T_{DIV}$ is set, the more generations without function evaluations for the changed operators are required.

For DE algorithms with linear population size reduction, further consideration is required.
In these algorithms, population size decreases linearly.
When diversity is improved by individuals redistribution, population size needs increase again.
Detailed steps are given below.
The value of population size is recorded when diversity is for the first time lower than $T_{DIV}$ in an execution.
When the course for diversifying begin, not only trial vectors but also target vectors are selected to increase population size until population size recovers to the recorded value.

\section{Experiments}
The ten DE algorithms are involved in experiments.
Their parameters excepting $MFES$ are from literature.
Three versions of the algorithms, original version (OV), version based on complete restart (CRV), and version based on our individuals redistribution (IRV), are compared.
Entry criterion is set the same for CRV and IRV. The reason why studying CRV is as below.
Complete restart in execution gives a new chance of evolution from beginning again. 
Now that population of DE is easy to fall into a local optimum, provided that the entry criterion of restart is set well, outcome of an execution of CRV setting a large value in $MFES$ corresponds to the best outcome of multiple executions of OV with standard $MFES$ setting.
In brief, although complete restart does not belong to a contribution in theory of DE, it may lead to improvement of solution in long execution.
If results based on CRV are better than results of OV, we can judge that parameters have been set well for IRV because IRV and CRV are based on the same entry criterion.
Furthermore, comparing results based on IRV with results based on CRV can further validate our finding.
%
Before formal experiments, we run OV of the ten DE algorithms for the CEC 2017 benchmark test suite.
$MFES$ is set $3.0E+06$, while dimensionality is set 30.
We do Friedman rank sum test at a 0.05 significance level for results and find that, at least for the CEC 2017 functions with 30 in dimensionality, DE algorithms selected from the competitions - L-SHADE-RSP, L-SHADE-EpSin, and L-SHADE - perform best among the ten algorithms.

In the first experiment, the three versions of the ten DE algorithms are compared based on the CEC 2017 benchmark test suite. 
Functions in the suite are set 30 in dimensionality.
Parameters for the three versions are listed in Table~\ref{table1}.
In the table, $MFES$ is required for all the three versions.
Moreover, $G_N$, $T_{IR}$, and $R$ are for both CRV and IRV, while $T_{DIV}$ and $T_{GEN}$ is for IRV only.  
\begin{table}[t]\scriptsize
\centering 
\begin{tabular}{lr|lr}
\toprule
Parameter                   & Value & Parameter & Value                              \\\midrule
{$MFES$}       & $6.0E+06$ for L-SHADE-RSP and   & $T_{GEN}$                 & 1000        \\
                            & $3.0E+06$ for the other algorithms &
$G_N$                        & 500                                 \\
{$T_{DIV}$ } & 1.0E-01, 5.0E-02, 1.0E-02, 5.0E-03, & $T_{IR}$                   & 1.0E-05                             \\
                      & 1.0E-03, 5.0E-04, 1.0E-04         
                      & $R$                          & 0.9 \\
\bottomrule                      
\end{tabular}
\caption{Parameters for the three versions in the first experiment
}
\label{table1}
\end{table}
We double $MFES$ for L-SHADE-RSP because,
under $MFES=3.0E+06$, restart can be hardly observed in CRV of the DE algorithm.
We employ seven values to threshold of diversity for returning to original mode - $T_{DIV}$ - because the suitable value of $T_{DIV}$ depends on not only algorithm but also fitness landscapes of function. 
Table~\ref{table2} gives a summary of results.
\begin{table}[t]\scriptsize
\centering 
\begin{tabular}{lrrr}
\toprule
              & CRV vs OV & IRV vs OV & IRV vs CRV \\\midrule
L-SHADE       & 15:1                & 16:2                        & 12:3                       \\
CoBiDE        & 23:0                & 23:0                        & 9:0                        \\
L-SHADE-EpSin & 4:1                 & 18:2                        & 14:3                       \\
SaDE/Mexp     & 19:1                & 19:1                        & 9:5                        \\
MPEDE         & 16:2                & 22:2                        & 8:2                        \\
ETI-JADE      & 23:1                & 25:0                        & 11:2                       \\
L-SHADE-RSP   & 4:3                 & 17:2                        & 17:4                       \\
EDEV          & 23:3                & 21:1                        & 11:8                       \\
MLCC-SI       & 19:3                & 18:3                        & 7:9                        \\
NDE           & 18:4                & 19:1                        & 11:2         \\    
\bottomrule     
\end{tabular}
\caption{Result summary of the first experiment. 
Former item and latter item of ratio denote times of significant improvement and statistical deterioration in terms of Wilcoxon's rank sum test at a 0.05 significance level, respectively}
\label{table2}
\end{table}

In addition, the ratio of times of showing the best result under different $T_{DIV}$ is given in Table~\ref{table3}.
\begin{table}[b]\scriptsize
\begin{tabular}{lr|lr}
\toprule
              Algorithm& Ratio&Algorithm& Ratio\\
              \midrule
L-SHADE       & 15:10:9:8:7:7:7 &ETI-JADE      & 10:10:6:6:6:4:7                                \\
CoBiDE        & 11:11:9:8:6:6:6  &L-SHADE-RSP   & 19:9:11:8:6:6:7                                                     \\
L-SHADE-EpSin & 13:13:9:7:6:6:6   &EDEV          & 5:6:7:5:3:5:8                                                    \\
SaDE/Mexp     & 14:10:7:7:5:2:3   &MLCC-SI       & 1:3:2:3:4:8:16                                                              \\
MPEDE         & 16:8:9:7:6:6:8     &NDE           & 10:12:10:6:4:4:3  \\\bottomrule                                 
\end{tabular}
\caption{Ratio of times of showing the best result under different $T_{DIV}$. Provided that the best result is obtained under more than one $T_{DIV}$ simultaneously, more than one positions are all added 1}
\label{table3}
\end{table}

It can be seen from Table~\ref{table2} that, for all the ten DE algorithms, CRV performs better than OV.
Nevertheless, in the comparison for L-SHADE-EpSin and L-SHADE-RSP, the advantage is just slight.
In fact, it can be observed from detailed outcomes that, within the current value of MFES, execution of OV of the two algorithms still do not completely fall into a local optimum in many cases.
CRV cannot show its best in this situation.
In short, according to Table~\ref{table2}, complete restart can improve results of DE algorithms as long as parameters for complete restart are set appropriately.

It can be seen from Table~\ref{table2} that IRV outperforms OV.
Further, comparison between IRV and OV show a larger ratio than comparison between CRV and OV for L-SHADE-EpSin, MPEDE, ETI-JADE, L-SHADE-RSP, EDEV, and NDE.
Meanwhile, the ratio is same in the comparison for CoBiDE, SaDE/Mexp.
Only in the comparison for L-SHADE and MLCC-SI, the ratio decreases.
More importantly, in the comparison for all the DE algorithms excepting MLCC-SI,
the times of IRV winning CRV is larger than the times of losing.
However, for EDEV, the advantage of IRV is not distinct.
In short, according to Table~\ref{table2}, IRV shows better performance than CRV for most of the algorithms.

According to Table~\ref{table3}, the best value of $T_{DIV}$ among the seven values is often large.
In detail, the best value for L-SHADE, CoBiDE, L-SHADE-EpSin, SaDE/Mexp, MPEDE, ETI-JADE, L-SHADE-RSP is 1.0E-01.
Meanwhile, 5.0E-02 is the best choice for CoBiDE, L-SHADE-EpSin, ETI-JADE, NDE.
It can be seen that, both 1.0E-01 and 5.0E-02 are the best choices for CoBiDE, L-SHADE-EpSin, and ETI-JADE.
Just EDEV and MLCC-SI are not fit for large value in $T_{DIV}$.

In the second experiment, the three versions are compared based on the CEC 2014 benchmark test suite.
Here, a part of the ten DE algorithms are selected to be executed.
Above all, L-SHADE-EpSin, one of the top performers in the CEC competitions and NDE, one of the latest DE algorithms, are the reasonable choices. 
Besides, ETI-JADE, which is based on partial restart, need be further studied.
Finally, the three ensemble DE algorithms - MPEDE, EDEV, and MLCC-SI - are involved in this experiment because the previous experiment shows that our scheme performs not so well for EDEV and is not appropriate to MLCC-SI at all.
Parameters for the three versions different with the previous experiment are listed in Table~\ref{table4}.
\begin{table}[t]\scriptsize
\centering
\begin{tabular}{lr}
\toprule
Parameter                   & Value                          \\ \midrule
$MFES$      & $6.0E+06$ for L-SHADE-EpSin
\\
$G_N$      
                            &  1000 for L-SHADE-EpSin \\
                            \bottomrule
\end{tabular}
\caption{Changed parameters for the three versions in the second experiment
}
\label{table4}
\end{table}
In fact, when $MFES$ and $G_N$ are set $3.0E+06$ and 500, respectively, for L-SHADE-EpSin, we find that complete restart leads to worse solutions in most cases.
It means that restart is called when execution still does not fall into a local optimum.
Therefore, we double $G_N$.
Meanwhile, we double $MFES$.
Table~\ref{table5} summarizes results.
\begin{table}[b]\scriptsize
\centering 
\begin{tabular}{lrrr}
\toprule
              Algorithms & CRV vs OV & IRV vs OV & IRV vs CRV  \\\midrule
L-SHADE-EpSin & 0:3     & 12:6     & 12:6     \\
MPEDE & 15:6	&18:4	&9:4\\
ETI-JADE      & 16:3 & 20:0 & 15:2 \\
EDEV          & 20:1 & 20:0 & 10:5  \\
MLCC-SI       & 16:6 & 9:8  & 6:7  \\
NDE           &14:6	 &17:3	&13:1\\
\bottomrule
\end{tabular}
\caption{Result summary of the second experiment}
\label{table5}
\end{table}

Then, ratio of times of showing the best result under different $T_{DIV}$ is given in Table~\ref{table6}.
\begin{table}[t]\scriptsize
\centering 

\begin{tabular}{lr|lr}
\toprule
              Algorithm& Ratio&Algorithm& Ratio
              \\\midrule
L-SHADE-EpSin & 11:11:11:8:12:13:8 & EDEV          &  9:5:9:8:7:2:3 \\      MPEDE &   13:6:11:5:4:5:6     &MLCC-SI       & 4:4:3:6:6:6:13\\
     ETI-JADE      & 9:3:7:7:9:9:12&NDE           &    8:9:6:8:4:6:7                               \\\bottomrule
\end{tabular}
\caption{Ratio of times of showing the best result under different $T_{DIV}$}
\label{table6}
\end{table}

It can be seen from Table~\ref{table5} that, for MPEDE, ETI-JADE, EDEV, MLCC-SI, and NDE, CRV outperforms OV.
However, for L-SHADE-EpSin, although the parameters are doubled, complete restart still cannot lead to improvement in any case but bring deterioration in cases.
According to the table, IRV always outperforms OV.
The advantage for MLCC-SI is very slight. Meanwhile, IRV wins CRV for the algorithms excepting MLCC-SI.
In addition, the advantage for EDEV and L-SHADE-EpSin is less than for MPEDE, ETI-JADE, and NDE.
It needs be emphasized that, for L-SHADE-EpSin, CRV cannot win OV.
That is, the parameters are still not for both complete restart and individuals redistribution.
However, IRV shows advantage on both OV and CRV.

According to Table~\ref{table6}, the best value of $T_{div}$ becomes smaller for L-SHADE-EpSin, MPEDE, ETI-JADE, and NDE than in the previous experiment.
Meanwhile, EDEV and MLCC-SI are still fit for the smallest value in $T_{div}$.

The summary of our experiments is given below.
Firstly, complete restart can improve DE algorithms provided that parameters are well set.
For example, if we set further larger value in $G_N$ and $MFES$, for CEC 2014 benchmark test suite, CRV of L-SHADE-EpSin may win OV of the algorithm.
CRV of ETI-JADE outperforming OV of the algorithm proves that complete restart and partial restart, which are very different in effects on evolution, can be used together in one DE algorithm.
More importantly, individuals redistribution can perform better than complete restart for most of the DE algorithms.
Especially, results of L-SHADE-EpSin for the CEC 2014 functions show that
individuals redistribution may even work well when the same entry criterion is not suitable for complete restart.
Nevertheless, a suitable value of $T_{div}$ varies for different algorithms and different functions.

Further, although even one of the best performer in the competitions among population-based metaheuristics - L-SHADE-EpSin and one of the latest DE algorithms - NDE are improved by us, our scheme leads to distinct effect for the three ensemble DE algorithms.
Although, for MPEDE, IRV show better performance than OV and CRV, our scheme performs not so well for EDEV and is not appropriate to MLCC-SI at all.
Our explanation is that, the more complicated an ensemble DE is, the worse the current scheme for calling individuals redistribution performs. 

\section{Conclusion and Future Work}
We find that the tendency of falling into local optimum widely exists in metaheuristics for real parameter single objective optimization.
In this paper, we proposed DE with individuals redistribution for keeping search ability in long run.
Our experiments based on ten DE algorithms and two benchmark test suites show that individuals redistribution is better at improving solutions for real parameter single objective optimization than complete restart.
Considering the leading position of DE in the field of real parameter single objective optimization, our finding contributes to the whole field.


Besides special scheme for calling individuals redistribution in ensemble DE algorithm, we want to obtain an adaptive method for setting $T_{DIV}$ to simplify application of individuals redistribution.
Moreover, after adaption, individuals redistribution may be realized in metaheuristics for parameter single objective optimization other than DE, such as HS-ES and UMOEA-\uppercase\expandafter{\romannumeral2}.
The above all will be accumulated as our future work. 
\bibliographystyle{named}
\bibliography{ijcai20}

\end{document}